\documentclass[preprint,12pt]{elsarticle}

\usepackage{amssymb}
\usepackage{amsmath}
\PassOptionsToPackage{hyphens}{url}\usepackage{hyperref}

\usepackage{listings}
\usepackage{xcolor}
\usepackage[doublespacing]{setspace}
\usepackage{booktabs}
\usepackage{multirow}
\usepackage{makecell}
\usepackage[nameinlink]{cleveref}

\definecolor{codegreen}{rgb}{0,0.6,0}
\definecolor{codegray}{rgb}{0.5,0.5,0.5}
\definecolor{codepurple}{rgb}{0.58,0,0.82}
\definecolor{backcolour}{rgb}{0.97,0.97,0.97}

\lstdefinestyle{python}{
    backgroundcolor=\color{backcolour},
    commentstyle=\color{codegreen},
    keywordstyle=\color{blue},
    numberstyle=\tiny\color{codegray},
    stringstyle=\color{codepurple},
    basicstyle=\ttfamily\footnotesize,
    breakatwhitespace=false,
    breaklines=true,
    captionpos=b,
    keepspaces=true,
    numbers=left,
    numbersep=5pt,
    showspaces=false,
    showstringspaces=false,
    showtabs=false,
    tabsize=2
}
\usepackage{lastpage}

\newcommand{\betac}{Beta}
\newcommand{\binomc}{Binomial}
\newcommand{\clm}{CLM}
\newcommand{\clmwk}{CLMWK}
\newcommand{\expc}{Exponential}
\newcommand{\drop}{Hybrid dropout}
\newcommand{\ecoc}{OBDECOC}
\newcommand{\sbc}{SB}
\newcommand{\triang}{Triangular}

\newcommand{\rev}[1]{{\color{black}#1}}

\journal{Neurocomputing}

\begin{document}

\begin{frontmatter}

\title{\texttt{dlordinal}: a Python package for deep ordinal classification}

\author[UCO]{Francisco Bérchez-Moreno} 
\author[IMIBIC]{Rafael Ayllón-Gavilán}
\author[UCO]{Víctor M. Vargas\corref{cor1}}
\author[UCO]{David Guijo-Rubio}
\author[UCO]{César Hervás-Martínez}
\author[UCO]{Juan C. Fernández}
\author[UCO]{Pedro A. Gutiérrez}

\cortext[cor1]{Corresponding author\\ E-mail address: \url{vvargas@uco.es}}

\address[UCO]{Department of Computer Science and Numerical Analysis, University of Córdoba, Córdoba, Spain}

\address[IMIBIC]{Department of Clinical-Epidemiological Research in Primary Care, Instituto Maimónides de Investigación Biomédica de Córdoba, Spain}

\begin{abstract}
\texttt{dlordinal} is a new Python library that unifies many recent deep ordinal classification methodologies available in the literature. Developed using PyTorch as underlying framework, it implements the top performing state-of-the-art deep learning techniques for ordinal classification problems. Ordinal approaches are designed to leverage the ordering information present in the target variable. Specifically, it includes loss functions, various output layers, dropout techniques, soft labelling methodologies, and other classification strategies, all of which are appropriately designed to incorporate the ordinal information. Furthermore, as the performance metrics to assess novel proposals in ordinal classification depend on the distance between target and predicted classes in the ordinal scale, suitable ordinal evaluation metrics are also included. \texttt{dlordinal} is distributed under the BSD-3-Clause license and is available at \url{https://github.com/ayrna/dlordinal}.
\end{abstract}

\begin{highlights}
\item Python package with ordinal classification methodologies for deep learning.
\item Ordinal dropout methodologies, output layers, soft labelling, losses and wrappers.
\item Easy integration with third party packages like \texttt{Skorch}.
\item Extensive test coverage.
\item User-friendly interface with intuitive design and comprehensive tutorials.
\end{highlights}

\begin{keyword}
Deep learning \sep Ordinal classification \sep Ordinal regression \sep Python \sep \texttt{PyTorch} \sep Soft labelling
\end{keyword}

\end{frontmatter}

\section{Introduction}

The term ordinal classification (also known as ordinal regression) refers to supervised learning problems characterised by an inherent order relationship between the class labels \cite{gutierrez2016current, perez2014classification}. In real-world scenarios, numerous problems of this nature arise. Specifically, in the healthcare domain, classification problems in which the categories follow a natural order are quite common \cite{tang2023disease}. One example from this domain is classifying the nature of a tumour lesion into varying degrees of severity \cite{arya2021multi,umamaheswari2024cnn}. In this case, the possible labels might be \textit{benign}, \textit{premalignant} (precancerous), \textit{in situ}, and \textit{malignant} (cancerous). Another example is grading the different levels of diabetic retinopathy \cite{pascual2024multivariate}, a complication of diabetes that affects vision. In this instance, the disease is often classified into stages such as \textit{no DR}, \textit{mild}, \textit{moderate}, \textit{severe}, and \textit{proliferative DR}. These examples illustrate two ordinal classification problems, as they involve distinct degrees of a condition and similarities between adjacent categories. Therefore, the primary objective of ordinal classification is to minimise the magnitude of misclassification errors, i.e. the distance between the target and predicted class on the ordinal scale.

This is accomplished by considering the varying costs associated with different types of errors. Specifically, classification errors between adjacent classes are less significant than errors between distant classes. Thus, in the former example, the error committed when misclassifying a \textit{benign} tumour as \textit{malignant} should be greater than the error incurred when misclassifying it as \textit{premalignant} (precancerous). Therefore, ordinal classification approaches are designed to account for the severity of misclassification errors by incorporating the ordinal nature of the labels into both the learning algorithm and the evaluation metrics.

Some existing packages, such as \texttt{ORCA} \cite{sanchez2019orca}, in Matlab, or \texttt{mord} \cite{pedregosa2015feature}, in Python, implement ordinal classification approaches for machine learning. Moreover, some previous works have released their source code\footnote{\url{https://github.com/rpmcruz/unimodal-ordinal-regression}}, as is the case of some unimodal distributions for deep ordinal classification tasks \cite{cardoso2023unimodal}. Nevertheless, these works do not provide a structured Python package, so they can not be easily installed and used. In this way, to the best of the authors' knowledge, there is not a framework that bridges the gap between ordinal classification and Deep Learning (DL), providing the researchers with a single framework integrating numerous ordinal approaches and other functionalities, useful for developing novel methodologies.

In this paper, we introduce \texttt{dlordinal}, a Python toolkit that implements ordinal classification methodologies for DL developed on top of the open-source \texttt{PyTorch} framework. Our software offers a wide number of DL approaches that leverage ordinal information inherent in the problem domain. The documentation and code of the \texttt{dlordinal} toolkit is available at \url{https://github.com/ayrna/dlordinal}.

\section{Code design and features}

The \texttt{dlordinal} package is designed with a modular architecture, facilitating its use and extension. It perfectly integrates with third-party software such as \texttt{skorch} \cite{skorch} for a wide range of classification tasks. The dependency level is kept to a minimum, relying on essential libraries including \texttt{scikit-learn} \cite{pedregosa2011scikit}, \texttt{numpy} \cite{harris2020array}, \texttt{PyTorch} \cite{paszke2019pytorch}, \texttt{pandas} \cite{mckinney2010data}, \texttt{scipy} \cite{virtanen2020scipy}, \texttt{scikit-image} \cite{van2014scikit}, \texttt{tqdm} \cite{da2019tqdm}, and \texttt{pillow}. Additionally, \texttt{dlordinal} supports downloading and loading common ordinal image benchmark datasets, such as FGNet \cite{fu2015robust} or Adience \cite{eidinger2014age}. This facilitates easy benchmarking of novel ordinal approaches.

\begin{figure}[!ht]
    \centering
    \includegraphics[width=\linewidth]{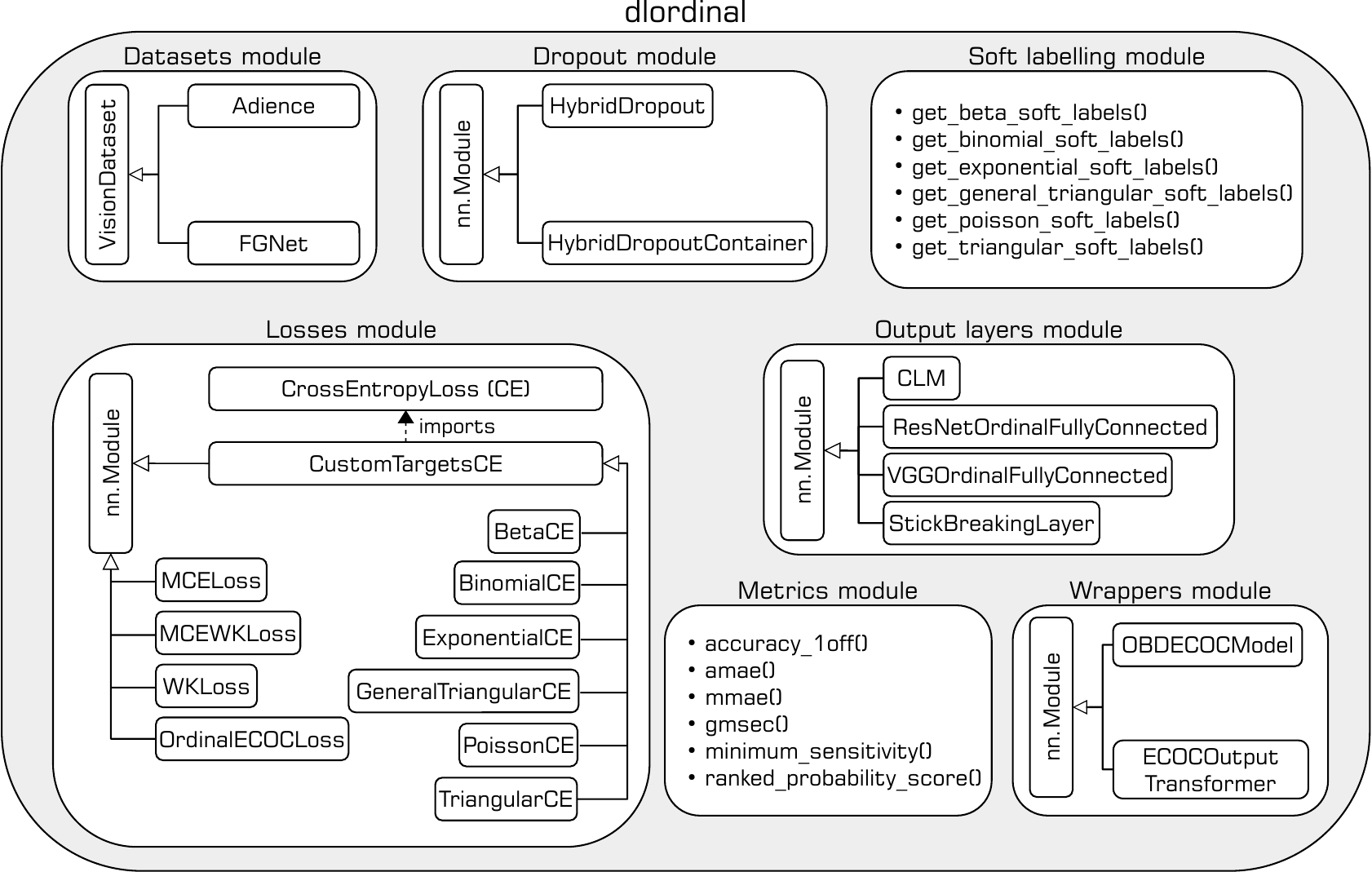}
    \caption{Representation of the classes and functions included in the various modules of the \texttt{dlordinal} package.}
    \label{fig:modules}
\end{figure}

As mentioned above, the \texttt{dlordinal} package features a modular design (see Figure \ref{fig:modules}), divided into several clearly defined modules:
\begin{itemize}
    \item \textbf{Datasets}: this module provides methods for loading the FGNet \cite{panis2016overview} and Adience \cite{eidinger2014age}, two ordinal datasets for benchmarking purposes. These datasets are useful for evaluating the performance of models in facial age estimation, a common ordinal classification task. FGNet offers a comprehensive collection of facial images across different ages, providing a valuable resource for studying facial ageing. Similarly, the Adience dataset includes unfiltered images collected from various sources, making it suitable for robust age and gender estimation. The tutorials included in \texttt{dlordinal} demonstrate how to effectively utilise these datasets for model training and evaluation.
    
    \item \textbf{Dropout}: this module includes a new hybrid dropout methodology \cite{berchez2024fusion} modifying the standard dropout approach \cite{Srivastava2014} to consider the order information of the target. In this way, the conventional dropout technique randomly drops neurons during training to prevent overfitting and improve generalisation, while the ordinal version analyses correlation between neurons and targets to promote those neurons that contribute to a better ordering of the classes. 
    
    \item \textbf{Output layers}: \texttt{dlordinal} implements the output layers proposed in the current literature. It includes an implementation of the Cumulative Link Models (CLM) output layer \cite{vargas2020clm}, which uses a projection obtained by an arbitrary deep neural network model. It also includes an implementation of the stick-breaking strategy proposed in \cite{liu2020unimodal}, which computes the class probabilities considering the ordinal information. Lastly, it implements an ordinal fully connected layer based on the Ordinal Binary Decomposition (OBD) approach described in \cite{barbero2023error}, which decompose the ordinal problem in several binary classification problems while preserving the ordinal information.

    \item \textbf{Soft labelling}: this module implements several functions for generating soft labels using different probability distributions \cite{vargas2024ebano}. This module forms the basis of the soft labelling strategy and is used in the corresponding loss functions. Soft labelling approaches replace one-hot encoded targets with the objective of regularising model predictions. These labels mitigate the effect of noisy labels \cite{liu2024distributed} in ordinal classification problems and lead to better generalisation performance. The underlying distributions generating these soft labels should be unimodal when dealing with ordinal classification. Specifically, for an ordinal classification problem with a given number of categories, the module includes state-of-the-art methodologies for generating soft labels using several distributions, such as Poisson or binomial \cite{liu2020unimodal}, exponential \cite{vargas2023exponential}, beta \cite{vargas2022unimodal}, and triangular \cite{vargas2023soft, vargas2023generalised}.

    \item \textbf{Losses}: several state-of-the-art ordinal loss functions are implemented in \texttt{dlordinal}. For example, regularised categorical cross-entropy loss functions based on the soft labelling strategy, considering any of the functions mentioned in the previous module. The weighted kappa loss function proposed in \cite{de2018weighted} is also included. The latter is an ordinal loss function based on a continuous version of the weighted kappa metric. Moreover, for the OBD approach \cite{barbero2023error}, a specific loss function is also implemented. This loss function considers the error of all the binary problems by computing the Mean Square Error (MSE) loss between the output of the model and the ideal output vector for each class.
    
    \item \textbf{Wrappers}: this module provides wrappers that enable the adaptation of any existing methodology to ordinal classification. An example can be found for the method introduced in \cite{barbero2023error}. It integrates any type of architecture into an OBD framework, and enhances the learning process for ordinal problems utilising the Ordinal Error Correcting Output Codes (ECOC) distance loss \cite{barbero2023error}.
    
    \item \textbf{Metrics}: in this module, several ordinal classification metrics are included: the 1-off accuracy, which measures the proportion of predictions that are either correct or off by one category. The Average Mean Absolute Error (AMAE) \cite{baccianella2009evaluation} and the Maximum Mean Absolute Error (MMAE) \cite{cruz2014metrics}. Both are robust measures that consider the imbalance degree present in the problem by averaging mean absolute errors or taking the maximum of the mean absolute errors computed for each class, respectively. The Quadratic Weighted Kappa (QWK), based on the weighted kappa index, aims to quantify the inter-rater agreement on classifying elements into a set of ordered categories. The QWK is specially appropriate to measure how well the model is predicting extreme classes \cite{de2018weighted}. The Ranked Probability Score (RPS), presented in \cite{epstein1969scoring}, measures the discrepancy between the predicted cumulative distribution and the actual cumulative distribution of the ordinal target. Finally, the Geometric Mean of the Sensitivities of the Extreme Classes (GMSEC) measures classification performance by focusing solely on the extreme classes, which are often the most important in certain ordinal problems.
\end{itemize}

\section{Sample code}

An example demonstrating the usage of some of the ordinal elements of the \texttt{dlordinal} toolkit, including a dataset, a loss function and some metrics, is as follows:

\begin{lstlisting}[language=Python]
import numpy as np
from dlordinal.datasets import FGNet
from dlordinal.losses import TriangularCrossEntropyLoss
from dlordinal.metrics import amae, mmae
from skorch import NeuralNetClassifier
from torch import nn
from torch.optim import Adam
from torchvision import models
from torchvision.transforms import Compose, ToTensor

# Download the FGNet dataset
fgnet_train = FGNet(
    root="./datasets",
    train=True,
    target_transform=np.array,
    transform=Compose([ToTensor()]),
)
fgnet_test = FGNet(
    root="./datasets",
    train=False,
    target_transform=np.array,
    transform=Compose([ToTensor()]),
)

num_classes_fgnet = len(fgnet_train.classes)

# Model
model = models.resnet18(weights="IMAGENET1K_V1")
model.fc = nn.Linear(model.fc.in_features, num_classes_fgnet)

# Loss function
loss_fn = TriangularCrossEntropyLoss(num_classes=num_classes_fgnet)

# Skorch estimator
estimator = NeuralNetClassifier(
    module=model,
    criterion=loss_fn,
    optimizer=Adam,
    lr=1e-3,
    max_epochs=25,
)

estimator.fit(X=fgnet_train, y=fgnet_train.targets)
train_probs = estimator.predict_proba(fgnet_train)
test_probs = estimator.predict_proba(fgnet_test)

# Metrics
amae_metric = amae(np.array(fgnet_test.targets), test_probs)
mmae_metric = mmae(np.array(fgnet_test.targets), test_probs)
print(f"Test AMAE: {amae_metric}, Test MMAE: {mmae_metric}")
\end{lstlisting}

\rev{
\section{Experiments and results}
This section outlines the experimental settings adopted to provide a benchmark of the main methodologies included in \texttt{dlordinal}.

\subsection{Methodologies}\label{Methodologies}
Given that the ordinal methodologies implemented in \texttt{dlordinal} are designed to be built on top of a base architecture, to enable a fair comparison, the same base architecture is selected. In this case, we consider the ResNet18 \cite{zhao2022deep} model pre-trained with ImageNet \cite{ImageNet8978153}. By default, the model implements a softmax output layer and is optimised using the categorical cross-entropy loss function.

The ordinal methodologies listed in \Cref{tab:benchmarked_methods} have been selected for this benchmark. This table shows the name of the methodology along with its type and location within the \texttt{dlordinal} package.

\begin{table}[!ht]
    \centering
    \rev{
    \caption{Methodologies considered for the experiments.}
    \label{tab:benchmarked_methods}
    \vspace{0.3cm}
    {\renewcommand{\arraystretch}{2.0}%
    \resizebox{\linewidth}{!}{
    \begin{tabular}{lll}
        \toprule \toprule
        Methodology & Type & Location(s) \\
        \midrule
        \betac{} \cite{vargas2022unimodal} & Loss & \texttt{dlordinal.losses.BetaCrossEntropyLoss}\\
        \binomc{} \cite{liu2020unimodal} & Loss & \texttt{dlordinal.losses.BinomialCrossEntropyLoss}\\
        \clm{} \cite{vargas2020clm} & Output layer & \texttt{dlordinal.output\_layers.CLM}\\
        \clmwk{} \cite{vargas2020clm} & \makecell[l]{Output layer\\ + loss} & \makecell[l]{\texttt{dlordinal.output\_layers.CLM} \\ \texttt{dlordinal.losses.WKLoss}}\\
        \expc{} \cite{vargas2023exponential} & Loss & \texttt{dlordinal.losses.BinomialCrossEntropyLoss}\\
        \drop{} \cite{berchez2024fusion} & Dropout & \makecell[l]{\texttt{dlordinal.dropout.HybridDropout} \\ \texttt{dlordinal.dropout.HybridDropoutContainer}}\\
        \ecoc{} \cite{barbero2023error} & \makecell[l]{Wrapper\\ + loss} & \makecell[l]{\texttt{dlordinal.wrappers.OBDECOCModel} \\ \texttt{dlordinal.losses.OrdinalECOCDistanceLoss}}\\
        \sbc{} \cite{liu2020unimodal} & Output layer & \texttt{dlordinal.output\_layers.StickBreakingLayer}\\
        \triang{} \cite{vargas2023soft} & Loss & \texttt{dlordinal.losses.TriangularCrossEntropyLoss}\\
        \bottomrule \bottomrule
    \end{tabular}}
    }
    }
\end{table}

\subsection{Training and evaluation procedure}\label{experimental_setup}
For this benchmark, three well-known datasets are considered: 1) Adience\footnote{\rev{\url{https://talhassner.github.io/home/projects/Adience/Adience-data.html}}} \cite{eidinger2014age}, 2) FGNet\footnote{\rev{\url{https://yanweifu.github.io/FG\_NET\_data/index.html}}} \cite{fu2015robust}, and 3) Diabetic Retinopathy\footnote{\rev{\url{https://www.kaggle.com/c/diabetic-retinopathy-detection/data}}} \cite{wang2018diabetic}.

The cross-validation of hyperparameters is conducted using a randomised search strategy using the values shown in Table \ref{crossvalidation}, with a maximum of $15$ iterations. For those hyperparameters not included in that table, the default values in \texttt{dlordinal} are used. Each hyperparameter configuration follows a $3$-fold strategy. Note that a total of $20$ runs, using a different random seed, are considered for robustness and fairness.

For the training procedure, a batch size of $128$ is used. The models are trained for a maximum of $100$ epochs. In addition, an early stopping strategy is adopted so that the training is terminated if no improvement in the validation loss is observed after $40$ epochs. With respect to the data partitioning, an stratified $75\%-25\%$ train-test split is employed. From the $75\%$ designated for training, an stratified $15\%$ partition is reserved for validation purposes.

To evaluate the different models, four metrics are considered, including two ordinal metrics such as the Quadratic Weighted Kappa (QWK) \cite{de2018weighted} and the Mean Absolute Error (MAE) \cite{ayllon2023dictionary}, the Correct Classification Rate (CCR) \cite{rosati2022novel} as a nominal classification measure, and the execution time. Note that both QWK and CCR are maximisation metrics, whereas the MAE has to be minimised.

\begin{table}[!ht]
    \centering
    \label{tab:parameters}
    \rev{
    \caption{Considered range of values for tuning the hyperparameters of each methodology.}
    \label{crossvalidation}
    \vspace{0.3cm}
    \resizebox{\textwidth}{!}{
    \begin{tabular}{llc}
        \toprule \toprule
        Classifier(s) & Hyperparameter description & Range of values \\
        \midrule

        \multirow{2}{*}{\betac, \binomc} & Learning rate & $\{10^{-4},10^{-3},10^{-2}\}$ \\
        & Smoothing factor eta & $\{0.8,1.0\}$ \\ [0.3cm]

        \sbc, \ecoc, \drop & Learning rate & $\{10^{-4},10^{-3},10^{-2}\}$ \\ [0.3cm]

        \multirow{3}{*}{\expc} & Learning rate & $\{10^{-4},10^{-3},10^{-2}\}$ \\
        & Exponent & $\{1.0, 1.5, 2.0\}$ \\
        & Smoothing factor eta & $\{0.8,1.0\}$ \\ [0.3cm]

        \multirow{3}{*}{\triang} & Learning rate & $\{10^{-4},10^{-3},10^{-2}\}$ \\
        & Adjacent class probability & $\{0.01, 0.05, 0.10\}$ \\
        & Smoothing factor eta & $\{0.8,1.0\}$ \\ [0.3cm]

        \multirow{1}{*}{\clm} & Learning rate & $\{10^{-4},10^{-3},10^{-2}\}$ \\[0.3cm]

        \multirow{1}{*}{\clmwk} & Learning rate & $\{10^{-4},10^{-2},10^{-3}\}$ \\[0.3cm]
        
        \bottomrule \bottomrule
    \end{tabular}
    }
    }
\end{table}

\subsection{Results}
This section provides the results obtained from the experiments detailed in Section \ref{experimental_setup}. Table \ref{table:resultados} shows the results for QWK, MAE, CCR, and computational time, as the Mean and Standard Deviation (SD) of the $20$ runs (Mean\textsubscript{SD}). For the Adience dataset, \triang{} achieves excellent results in MAE ($0.335$) and CCR ($0.753$), whereas \betac{} is the best in QWK ($0.924$). On the other hand, the \drop{} method is the fastest approach ($74.01$ seconds), achieving competitive results. For the FGNet dataset, \clmwk{} achieves the best results in terms of QWK ($0.846$), while the \triang{} estimator again outperforms the rest of the approaches in MAE ($0.491$) and CCR ($0.594$). Similarly, \drop{} is the most efficient, with an execution time of $10.69$ seconds. For the Diabetic Retinopathy dataset, \clmwk{} emerges as the top-performing method for all the metrics: QWK ($0.624$), MAE ($0.415$) and CCR ($0.687$). In this case, \ecoc{} achieves the shortest execution time ($154.90$ seconds). Overall, these results suggest that performance varies depending on the characteristics of the dataset and the metrics considered (including the one used in the cross-validation stage), with methods such as \betac{}, \triang{} and \clmwk{} achieving strong classification performance. Meanwhile, the \drop{} method consistently stands out in terms of computational time across all datasets, except for Diabetic Retinopathy, where it is outperformed by \ecoc{}.

Overall, these results highlight the effectiveness of the proposed approaches in leveraging ordinal information while offering a trade-off between performance and computational efficiency.

\begin{table}[ht!]
    \centering
    \rev{
    \caption{Results for each metric in the test datasets, expressed as the Mean and Standard Deviation (SD) for the 20 runs, Mean\textsubscript{SD}. The best result for each metric is highlighted in \textbf{bold}, whereas the second best is in \textit{italics}.}
    \label{table:resultados}
    \vspace{0.3cm}
    \resizebox{\textwidth}{!}{
        \begin{tabular}{llcccc}
        \toprule \toprule
        Dataset & Estimator & QWK & MAE & CCR & Time (seg) \\
        \midrule
        \multirow{9}{*}{Adience} & \betac & \textbf{0.924}$_{0.016}$ & \textit{0.349}$_{0.080}$ & \textit{0.730}$_{0.060}$ & 289.68$_{117.22}$ \\
         & \binomc & 0.832$_{0.086}$ & 0.773$_{0.170}$ & 0.442$_{0.052}$ & 353.85$_{82.72}$ \\
         & \clm & 0.902$_{0.019}$ & 0.437$_{0.075}$ & 0.670$_{0.055}$ & 1587.81$_{322.26}$ \\
         & \clmwk & 0.902$_{0.029}$ & 0.484$_{0.128}$ & 0.617$_{0.087}$ & 1090.42$_{768.00}$ \\
         & \expc & 0.918$_{0.026}$ & 0.391$_{0.109}$ & 0.694$_{0.076}$ & 263.45$_{44.55}$ \\
         & \drop & 0.812$_{0.037}$ & 0.740$_{0.130}$ & 0.525$_{0.057}$ & \textbf{74.01}$_{20.09}$ \\
         & \ecoc & 0.916$_{0.015}$ & 0.376$_{0.077}$ & 0.722$_{0.056}$ & 1009.82$_{704.51}$ \\
         & \sbc & 0.884$_{0.032}$ & 0.441$_{0.105}$ & 0.701$_{0.065}$ & 874.35$_{730.67}$ \\
         & \triang & \textit{0.923}$_{0.019}$ & \textbf{0.335}$_{0.085}$ & \textbf{0.753}$_{0.058}$ & \textit{249.47}$_{140.76}$ \\
        \midrule
        \multirow{9}{*}{FGNet} & \betac & \textit{0.839}$_{0.034}$ & \textit{0.510}$_{0.088}$ & \textit{0.564}$_{0.050}$ & 47.56$_{28.45}$ \\
         & \binomc & 0.824$_{0.030}$ & 0.614$_{0.072}$ & 0.468$_{0.054}$ & 49.74$_{22.83}$ \\
         & \clm & 0.793$_{0.057}$ & 0.592$_{0.132}$ & 0.514$_{0.071}$ & 65.70$_{22.91}$ \\
         & \clmwk & \textbf{0.846}$_{0.027}$ & 0.534$_{0.090}$ & 0.520$_{0.076}$ & 66.49$_{26.48}$ \\
         & \expc & 0.829$_{0.046}$ & 0.551$_{0.081}$ & 0.523$_{0.052}$ & 52.49$_{31.08}$ \\
         & \drop & 0.706$_{0.160}$ & 0.763$_{0.177}$ & 0.463$_{0.055}$ & \textbf{10.69}$_{2.90}$ \\
         & \ecoc & 0.760$_{0.058}$ & 0.639$_{0.104}$ & 0.501$_{0.067}$ & \textit{20.03}$_{14.94}$ \\
         & \sbc & 0.760$_{0.076}$ & 0.603$_{0.107}$ & 0.517$_{0.063}$ & 29.31$_{21.61}$ \\
         & \triang & 0.828$_{0.036}$ & \textbf{0.491}$_{0.057}$ & \textbf{0.594}$_{0.037}$ & 29.09$_{20.44}$ \\
        \midrule
        \multirow{9}{*}{Retinopathy} & \betac & 0.374$_{0.038}$ & 1.046$_{0.087}$ & 0.094$_{0.006}$ & \textit{232.64}$_{61.20}$ \\
         & \binomc & 0.387$_{0.048}$ & 1.049$_{0.110}$ & 0.089$_{0.006}$ & 356.60$_{538.93}$ \\
         & \clm & \textit{0.541}$_{0.058}$ & \textit{0.528}$_{0.132}$ & \textit{0.634}$_{0.096}$ & 2904.94$_{2500.61}$ \\
         & \clmwk & \textbf{0.624}$_{0.023}$ & \textbf{0.415}$_{0.038}$ & \textbf{0.687}$_{0.039}$ & 2979.26$_{1374.41}$ \\
         & \expc & 0.408$_{0.041}$ & 1.004$_{0.082}$ & 0.092$_{0.005}$ & 1206.61$_{732.51}$ \\
         & \drop & 0.428$_{0.084}$ & 0.793$_{0.224}$ & 0.487$_{0.133}$ & 233.98$_{102.20}$ \\
         & \ecoc & 0.282$_{0.181}$ & 1.348$_{0.678}$ & 0.220$_{0.117}$ & \textbf{154.90}$_{78.84}$ \\
         & \sbc & 0.447$_{0.133}$ & 0.688$_{0.200}$ & 0.554$_{0.116}$ & 269.33$_{79.49}$ \\
         & \triang & 0.427$_{0.071}$ & 0.815$_{0.191}$ & 0.414$_{0.145}$ & 250.42$_{140.19}$ \\
        \bottomrule \bottomrule
        \end{tabular}
    }
    }
    \end{table}
}

\section{Conclusions}
\texttt{dlordinal} has been developed to bring the practitioners with a Python package including deep ordinal classification methodologies proposed in recent literature. It is built on top of the well-known \texttt{PyTorch} framework. Up-to-the-knowledge of the authors, there are no previous software packages unifying all deep ordinal classification approaches under a single framework. Consequently, the scientific community should benefit from using \texttt{dlordinal} to develop and test novel methodologies and to apply existing ones to real-world problems with minimal effort.

\section*{Acknowledgments}
The present work has been supported by the ``Agencia Estatal de Investigación (España)'', Spanish Ministry of Research and Innovation (grant ref.: PID2023-150663NB-C22 / AEI / 10.13039 / 501100011033), by the European Commission, project Test and Experiment Facilities for the Agri-Food Domain, AgriFoodTEF (grant ref.: DIGITAL-2022-CLOUD-AI-02, 101100622), and by the ENIA International Chair in Agriculture, University of Córdoba (grant ref.: TSI-100921-2023-3), funded by the Secretary of State for Digitalisation and Artificial Intelligence and by the European Union - Next Generation EU. Recovery, Transformation and Resilience Plan and by the University of Córdoba through competitive grants for Andalusian society challenges (grant ref.: PP2F\_L1\_15). F. Bérchez-Moreno has been supported by ``Plan Propio de Investigación Submodalidad 2.2 Contratos predoctorales'' of the University of Córdoba. R. Ayllón-Gavilán has been supported by the ``Instituto de Salud Carlos III'' (ISCIII) and EU (grant ref.: FI23/00163).

\bibliographystyle{elsarticle-harv}
\bibliography{bibliography}

\end{document}